\documentclass[10pt,twocolumn,letterpaper]{article}

\usepackage{iccv}              
\newcommand{\mycomment}[1]{}

\usepackage{graphicx}
\usepackage{amsmath}
\usepackage{amssymb}
\usepackage{booktabs}
\usepackage{float}

\usepackage{placeins}

\usepackage{booktabs}
\usepackage{mathtools}

\usepackage{xcolor,color,framed}
\usepackage{url, overpic, cases, contour}
\usepackage{multirow, makecell}
\usepackage{adjustbox}

\definecolor{iccvblue}{rgb}{0.21,0.49,0.74}
\usepackage[pagebackref,breaklinks,colorlinks,allcolors=iccvblue]{hyperref}

\usepackage[capitalize]{cleveref}

\usepackage{commath}

\usepackage{amsmath}
\usepackage{amssymb}
\usepackage{booktabs}

\usepackage{placeins}

\usepackage{xcolor,color,framed}
\usepackage{url, overpic, cases, contour}
\usepackage{multirow, makecell}

\definecolor{shadecolor}{rgb}{0.92,0.92,0.92}

\usepackage{adjustbox}
\makeatletter %
\@namedef{ver@everyshi.sty}{}
\makeatother
\usepackage{tikz}

\newcommand{\name}{0}
\newcommand{\h}{0}
\newcommand{\w}{0.15}
\newcommand{\wa}{0.15}
\newlength \g

\usepackage[pagebackref,breaklinks,colorlinks]{hyperref}
\usepackage[symbol]{footmisc}

\usepackage{tikz}
\newcommand*\circled[1]{\tikz[baseline=(char.base)]{
		\node[shape=circle,draw,inner sep=0.1pt] (char) {#1};}}

\usepackage[capitalize]{cleveref}
\crefname{section}{Sec.}{Secs.}
\Crefname{section}{Section}{Sections}
\Crefname{table}{Table}{Tables}
\crefname{table}{Tab.}{Tabs.}

\def\mC{{\mathcal C}}

\def\mff{{ F}}

\DeclareMathAlphabet\mathbfcal{OMS}{cmsy}{b}{n}

\def\0{{\bf 0}}
\def\1{{\bf 1}}

\def\mypapertitle{{JFFRA }}

\usepackage[capitalize]{cleveref}
\crefname{section}{Sec.}{Secs.}
\Crefname{section}{Section}{Sections}
\Crefname{table}{Table}{Tables}
\crefname{table}{Tab.}{Tabs.}
\def\paperID{9960} 
\def\confName{ICCV}
\def\confYear{2025}

\title{Joint Flow And Feature Refinement Using  Attention For Video Restoration}

\author{Ranjth Merugu$^{2}$\\
{\tt\small ranjith.merugu@stonybrook.edu}
\and
Mohammad Sameer Suhail$^\text{* 1}$\\
{\tt\small mo.suhail@samsung.com}
\and
Akshay P Sarashetti$^\text{* 1}$\\
{\tt\small akshay.p@samsung.com}
\and
Venkata Bharath Reddy Reddem$^\text{* 1}$\\
{\tt\small r.reddy@samsung.com}
\and
Pankaj Kumar Bajpai$^{1}$\\
{\tt\small pankaj.b@samsung.com}
\and
Amit Satish Unde$^{1}$\\
{\tt\small amit.unde@samsung.com}
}

\begin{document}
\maketitle
\def\thefootnote{*}\footnotetext{Equal contribution}\def\thefootnote{\arabic{footnote}}
\def\thefootnote{1}\footnotetext{Samsung R\&D Institute India, Bangalore}\def\thefootnote{\arabic{footnote}}
\def\thefootnote{2}\footnotetext{Stony Brook University}\def\thefootnote{\arabic{footnote}}

\begin{abstract}

Recent advancements in video restoration have focused on recovering high-quality video frames from low-quality inputs. Compared with static images, the performance of video restoration significantly depends on efficient exploitation of temporal correlations among successive video frames. The numerous techniques make use of  temporal information via flow-based strategies or recurrent architectures. However, these methods often encounter difficulties in preserving temporal consistency as they utilize degraded input video frames. To resolve this issue, we propose a novel video restoration framework named Joint Flow and Feature Refinement using Attention (JFFRA). The proposed JFFRA  is based on key philosophy of iteratively enhancing data through the synergistic collaboration of flow (alignment) and restoration. By leveraging previously enhanced features to refine flow and vice versa, JFFRA enables efficient feature enhancement using temporal information. This interplay between flow and restoration is executed at multiple scales, reducing the dependence on precise flow estimation. Moreover, we incorporate an occlusion-aware temporal loss function to enhance the network's capability in eliminating flickering artifacts. Comprehensive experiments validate the versatility of JFFRA across various restoration tasks such as denoising, deblurring, and super-resolution. Our method demonstrates a remarkable performance improvement of up to 1.62 dB compared to state-of-the-art approaches.
\end{abstract}    

\section{Introduction}
\label{sec:intro}
With advances in smart phones cameras and its widespread use, user-generated content in the form of images/videos is exponentially growing. However, despite significant progress in camera sensor technology, videos captured using mobile cameras suffer from degradation in perceptual quality due to the presence of noise, blur and compression artifacts \cite{argaw2021restoration,kim2024towards}. The artifacts in captured videos are complex and stochastic in nature. It causes severe degradation, especially in unconstrained scenarios such as data acquisition in low-light environment, low-cost mobile devices and unwanted camera movement. Removing degradation is challenging, as these inverse problems are ill-posed \cite{sheth2021unsupervised,nah2019ntire}. Therefore, video restoration ~\cite{liang2022recurrent,liang2021swinir,li2021arvo,liang2024vrt,kim2024towards} which is the process of  bringing videos back to life by enhancing its quality is an important problem in the field of computer vision.


\begin{figure}[htbp]
    \centering
    \includegraphics[width=0.45\textwidth]{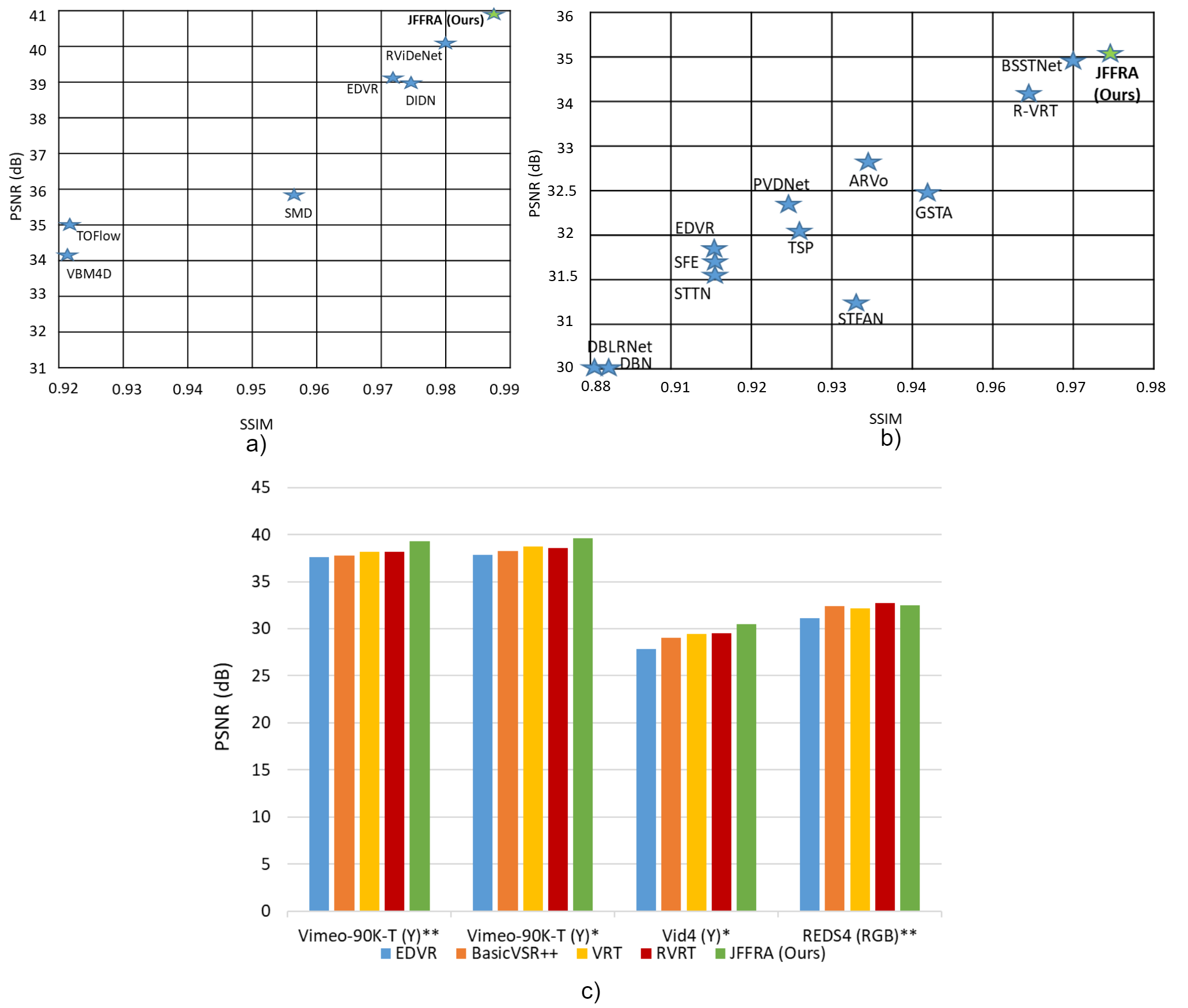}
    \caption{The proposed JFFRA framework achieves the state-of-the-art performance on various video restoration tasks such as a) Video Denoising, b) Video Deblurring, and c) Video Super-resolution. }
    \label{fig:comp_fig_3}
    \vspace{-5mm}
\end{figure}

Video restoration ~\cite{liang2024vrt,liang2022recurrent} is gaining an attention of research community owing to increased potential of storage and processing power of mobile devices and cloud computing. The key challenges in video restoration can be categorized into two streams as (1) exploiting temporal correlation between adjacent frames and (2) flicker removal. Over the past few years, variety of video restoration algorithms ~\cite{liang2024vrt,kim2024towards,truong2024self} ranging from classical methods ~\cite{cao2021vsrt,caballero2017real,wang2019edvr,tian2020tdan,su2017deep,li2021arvo,liang2022recurrent,liang2024vrt,kim2024towards} to emerging learning based approaches have been proposed in the literature. After an exhaustive survey, it is noticed that classical methods based on either frame averaging, spatio-temporal patch based processing or non-local-means filtering fail to model inter-frame temporal correlation effectively, resulting in sub-optimal performance. With the emergence of convolutional neural network (CNN), deep-CNN based algorithms for video restoration ~\cite{liang2024vrt,huang2015bidirectional,maggioni2012bm4d,maggioni2021efficient}has shown significant improvement in the performance and is the promising research direction.

The CNN-based methods exploit temporal redundancy through implicit or explicit motion modeling. The implicit motion modeling can be achieved through multi-frame processing, recurrent neural network or Long Short-term Memory (LSTM)\cite{yu2019review,staudemeyer2019understanding} which holds the potential to model spatio-temporal correlation between adjacent frames. Albeit promising, these methods lacks in fully exploiting the temporal redundancy between adjacent frames and can be prone to error accumulation. This problem of inadequate correlation exploitation can be alleviated through explicit motion modeling using optical flow ~\cite{fischer2015flownet,sun2018pwc,teed2020raft}.

The optical flow based restoration algorithms ~\cite{tassano2019dvdnet,xue2019toflow} first estimate motion vectors between frames. The temporal alignment across consecutive frames is accomplished through warping of pixels from adjacent frames with respect to a common reference frame. This alignment facilitates the restoration process through the exploitation of temporal coherence. It also ensures that degradation patterns are consistently addressed across the video frames. However, the warping operation can lead to troublesome ghosting problem, which may cause performance degradation. In addition, optical flow based methods also suffer from inevitable and challenging occlusion ~\cite{wang2018occlusion} problem, further damaging performance.

To overcome the drawbacks of optical flow, attention mechanisms ~\cite{vaswani2017attention,xia2022vision, cao2022datsr} have become pivotal, particularly for tasks that require spatial and temporal processing. In this direction, the current state-of-the-art networks such as video restoration transformer (VRT) ~\cite{liang2024vrt} leverages on mutual and self-attention mechanism to model long-range spatio-temporal dependency. While VRT demonstrated significant performance gains for different video restoration tasks, it has two major limitations. First, its long-range temporal information modeling ability is limited. This can be attributed to the use of window based attention mechanism for implicit motion estimation on image features. Hence, VRT fails to handle large motions efficiently. Second, the implicit motion modeling in feature space is reminiscent of learning  un-referenced functions which are difficult to optimize. Therefore, this implicit motion modeling can be more error-prone, particularly while modeling sufficiently moderate motions.  

In this paper, we address the problems associated with both implicit and explicit motion modeling through joint refinement of flow and frame features. We propose Joint Flow and Feature Refinement using Attention (JFFRA) network that handles three major problems: (1) it prevents errors in implicit motion modeling by utilizing initial flow computed on the input frames, (2) through iterative flow and feature refinement it continually corrects the explicit motion modeling, and (3) by utilizing a novel occlusion-aware temporal loss function it reduces the misalignment between restored frames, reducing the flickering problem in video restoration tasks.    

Our main contributions can be summarized as follows:
\begin{enumerate}
	\item We propose a novel Joint flow and feature refinement using Attention
	(JFFRA ) module that iteratively refines the flow and frame features in collaborative manner, resulting in continual improvement of motion modeling and restoration.
	\item We introduce a novel  occlusion-aware temporal loss function that handles the temporal alignment across restored frames and eliminate flickering artifacts.
	\item To the best of our knowledge, we are the first to demonstrate that iterative refinement with mutual reinforcement between flow (alignment) and restoration consistently outperforms the end-to-end learning method in terms of restored video frames quality.
	\item We validate the effectiveness and generality of proposed JFFRA approach on range of video restoration task such as denoising, superresolution and deblurring, achieving state-of-the-art (SOTA) performance and outperforming the existing methods by up to 1.62 dB on benchmark datasets as shown in \figurename ~\ref{fig:comp_fig_3}.
\end{enumerate}

\section{Related Work}
In this section, we briefly review related work on video restoration and optical flow.

\subsection{Video Restoration}
Early video-restoration methods leveraged recurrent networks to capture sequential information effectively. Recent advancements have seen improvements such as enhanced grid propagation ~\cite{8296826, 9150666},flow based alignment techniques and deformable convolutions  ~\cite{wang2019edvr,zhu2019dcnv2,cao2022datsr}. Some approaches have adopted asymmetric loss functions to optimize network performance, while others have proposed patch-based algorithms ~\cite{7448946, vaksman2021patch} that exploit correlations among patches for better restoration. Combining patch-based frameworks  ~\cite{vaksman2021patch,vaksman2021pacnet}with convolution neural networks has also been explored, leading to the development of more sophisticated patch-crafting techniques.
 Additionally, self-supervised learning techniques have been applied to handle unknown degradation effectively. Blind restoration methods ~\cite{claus2019videnn,zhang2022scunet} trained on diverse degradations  have also emerged. State-of-the-art techniques, including video restoration transformers ~\cite{liang2024vrt,chan2021basicvsrpp} with parallel frame prediction, have set new benchmarks in the field. However, existing optical-flow-based methods ~\cite{tassano2019dvdnet,xue2019toflow} often overlook the impact of degradation on flow estimation, which can result in incorrect feature alignment. While task-oriented flow learning approaches have been proposed, they sometimes fail to refine and compensate for inaccuracies in flow estimation comprehensively.

To address this issue, a novel JFFR  block is proposed that aligns features with flow values and refines flow using the corrected features. The alignment of features with the refined flow enhances the quality of video-restoration results.

\subsection{Optical Flow}\label{AA}
Optical flow estimation is crucial in any video enhancement task. The idea of optical flow was first proposed in Flownet \cite{fischer2015flownet}, recent advancements show that flow could be used in multiple tasks. Video restoration tasks \cite{maggioni2021efficient,tassano2020fastdvdnet,liang2024vrt} achieved state-of-the-art results utilizing flow for frame alignment. The flow computed on degraded frames can be corrupted. So, flow correction is necessary for video enhancements \cite{tassano2019dvdnet,maggioni2021efficient}.
To improve the efficiency of alignment, a novel refinement block is proposed in this paper that refines flow based on intermediate frame features and vice versa, both of which are progressively improvised over the original optical flow and degraded frames respectively. The refined flow is used as offset to guide the network for more coherent temporal alignment in intermediate frame features. These refined frame features are again used to correct the flow.

\begin{figure*}
	\centering
	\includegraphics[width=0.9\linewidth,height=0.6\linewidth]{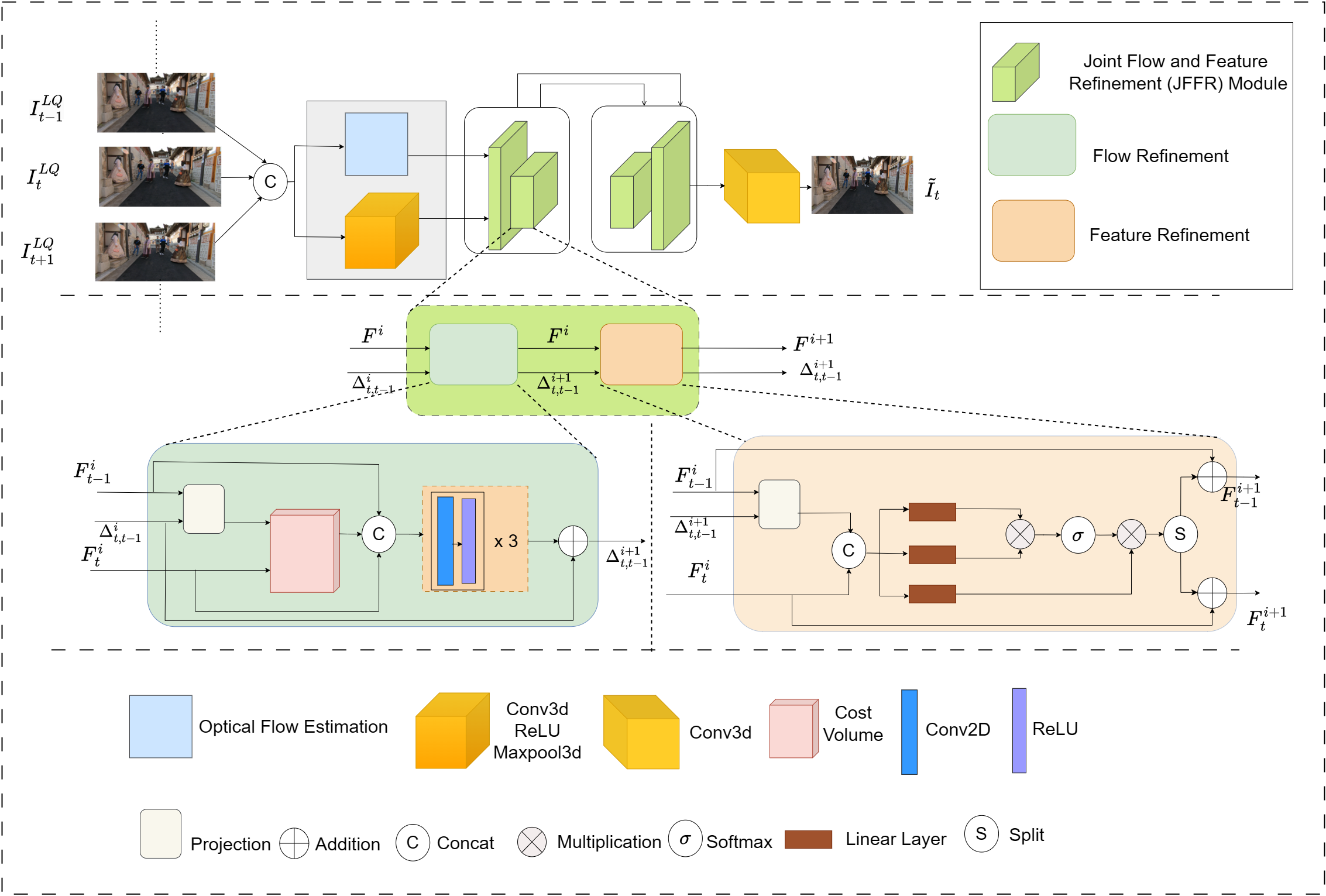}
	\caption{Architecture of the proposed joint flow and feature refinement using attention framework(JFFRA).}
	\label{fig:main_arch}
\end{figure*}

	
	\section{Proposed Methodology}
	\subsection{Overall Framework}
	Let $I^{\textit{LQ}}\in \mathbb{R}^{T\times H\times W\times C}$ be a sequence of degraded low-quality frames (LQ) input frames and  $I^{\textit{HQ}}\in \mathbb{R}^{T \times H\times W\times C}$ be a sequence of high-quality ground-truth frames (HQ). $T$, $H$, $W$, $C$ represents frame number, height, width, channel number respectively. We consider $2N+1$ consecutive frames as input when restoring a reference frame at time $t$, $I^{\textit{LQ}}_t$, and denote them as $[I^{\textit{LQ}}_{t-N},...,I^{\textit{LQ}}_{t+N}]$.  This process is repeated for every consecutive sequence of $2N+1$ frames using shifted window approach. As illustrated in ~\figurename$ $ \ref{fig:main_arch}, the proposed network can be primarily divided into two parts: (1) temporal context extraction, (2) joint flow and feature refinement module, which is trained on temporal loss with occlusion-aware masking. 
	
	\subsection{Integrating Temporal Context}
	The 3D convolutions are applied to the consecutive input frames to exploit the temporal context. For a consecutive sequence of $2N+1$ frames, we represent extracted spatio-temporal features as  $F^{\textit{i}}\in \mathbb{R}^{(2N+1)\times H^{'}\times W^{'}\times C^{'}}$,  where \(H'\), \(W'\) are the spatial dimensions and \(C^{'}\) is the number of output channels. In addition, temporal information between input frames is further exploited by explicit motion modeling through optical flow. To be more specific, the optical flow is computed using the pre-trained RAFT ~\cite{teed2020raft} model on $2N$ LQ frames with respect to the current frame, $I^{\textit{LQ}}_t$. The initial optical flow between $I^{\textit{LQ}}_t$ and $I^{\textit{LQ}}_{t-N}$ is denoted as $\Delta^{0}_{t, t-N}$.
	The temporally rich features obtained through 3D convolutions together with optical flow information are then passed to joint flow and feature refinement module.
	
	\subsection{Joint Flow and Feature Refinement Module}
	The initial optical flow is computed on LQ frames, its accuracy can be affected due to the presence of degradation in the input frames. The proposed JFFR module is focused on jointly correcting the initial flow and then  refining frame features in an alternating manner. 
	
	Without any loss in generality, the proposed method is described with a sequence of 3 consecutive frames $ (i.e., N = 3)$. Initially,  the spatio-temporal features ($i.e.,$ the input frames) $F^{\textit{0}} = [F^{\textit{0}}_{t-1}, F^{\textit{0}}_t, F^{\textit{0}}_{t+1}]$ and  flow ($i.e$., optical flow on input frames) $[\Delta^{0}_{t, t-1}, \Delta^{0}_{t, t+1}]$ are passed as input to the JFFR module. Specifically, given an input features and associated flow to the $i$-th level of encoder or decoder, the refined flow and features can be formulated as,
	\begin{equation}
		F^{\textit{i+1}}, \Delta^{i+1}_{t, t-1}, \Delta^{i+1}_{t, t+1}	= \text{JFFR}(F^{\textit{i}}, \Delta^{i}_{t, t-1}, \Delta^{i}_{t, t+1}).
	\end{equation}
	
	The proposed JFFR module first corrects the initial flow using cost volume and then  refine frame features using the corrected flow in an alternating manner. 
	
	\subsubsection{Cost volume computation} 
	The cost volume is computed independently on previous and next frame with respect to the reference frame. For the sake of simplicity, we present the cost volume computation with respect to previous frame only. The cost volume is computed in two steps. First, the features $F^{\textit{i}}_{t-1}$ are spatially transformed using the corresponding flow $\Delta^{i}_{t, t-1}$. It can be represented as,
	\begin{equation}
		F^{\textit{i}, {Proj}}_{t-1} = \mathcal{T}(F^{\textit{i}}_{t-1}, \Delta^{i}_{t, t-1}),
	\end{equation} 
	where $\mathcal{T}(.)$ is a function that projects or spatially transform the features using corresponding flow values. We use grid sampling to transform the features using flow values.  
	
	The cost volume is constructed to capture the $L_1$ distance between the features of the current frame and projected frame over a $9 \times 9$ search window. It is represented mathematically as,
	
	\begin{equation}
		\mathcal{C} = \mathcal{F}(F^{\textit{i}, {Proj}}_{t-1}, F^{\textit{i}}_{t}) \in \mathbb{R}^{d_h \times d_w \times H' \times W'},
	\end{equation}
	where $\mathcal{F}(.)$ is a function that transform 3D features into 4D cost volume. Here, \(d_h\) and \(d_w\) are the search dimensions in \(x\) and \(y\) directions, such that,
	\begin{equation}
		\mC[l,m,h,w] = \norm{\mff^{i}_{t}[h,w]-{\mff^{{i}, {Proj}}_{t-1}}[h+l,w+m]}_{1},
	\end{equation}
	where $l,m$ $\in$ $[-4, 4]$  based on the size of search window 9 $\times$ 9.

	The cost volume reflects the correlation between features of the current frame and associated features inside the search window of the projected frame. In order to produce the final result, we still need to decide maximum similarity matching within the search window for each feature of the reference frame. Therefore, the computed  cost volume $\mathcal{C}$ in conjunction with feature maps are processed using flow head network. The flow head refines flow values by utilizing the similarity information encoded in the cost volume. To ease the burden of feature learning, we employ offset flow learning and only predict deviation from the previously computed flow. The updated flow is given as,
	\begin{equation}
		\Delta^{i+1}_{t, t-1} = \Delta^{i}_{t, t-1} + \text{Flow Head}(\mC,\mff^{i}_{t}, \mff^{i}_{t-1}).
	\end{equation}

	\subsubsection{Feature refinement} 
	In this section, we integrate attention mechanism to refine features using the updated flow. Given a previous and next frame features $F_{t-1}^i$ and  $F_{t+1}^i$ respectively, the updated features are computed as:
	\begin{equation}
		F_{t-1}^{i, Update} = \mathcal{T}(F_{t-1}^i, \Delta^{i+1}_{t, t-1}),
	\end{equation} 
	
	\begin{equation}
		F_{t+1}^{i, Update} = \mathcal{T}(F_{t+1}^i, \Delta^{i+1}_{t, t+1}).
	\end{equation} 
	
	We then concatenate updated and reference frame features and compute the \emph{query} $Q^i$, \emph{key} $K^i$ and \emph{value} $V^i$ by linear projection as:
	\begin{equation}
		F_t^{i, Con} = F_{t}^i \;  \circled{C} \; F_{t-1}^{i, Update} \; \circled{C} \; F_{t+1}^{i, Update},
	\end{equation} 
	
	\begin{equation}
		Q_t^i=F_t^{i, Con}W^Q, \; K_t^i=F_t^{i, Con}W^K, \; V_t^i=F_t^{i, Con}W^V,
	\end{equation} where the operator  $\circled{C}$ represents concatenation and $W^Q$, $W^K$ and $W^V$ are learned projection matrices. We then compute attention map, $A_t^i$, using \emph{query} and \emph{key}. Finally, residuals are extracted as:
	\begin{equation}
		\mathcal{R}_t^i,\;\mathcal{R}_{t-1}^i, \; \mathcal{R}_{t+1}^i =  A_t^i V_t^i.
	\end{equation}
	The refined feature maps are computed as:
	\begin{equation}
		F_{t}^{i+1}, F_{t-1}^{i+1}, F_{t+1}^{i+1} = F_{t}^i + \mathcal{R}_t^i, F_{t-1}^i + \mathcal{R}_{t-1}^i, F_{t+1}^i+\mathcal{R}_{t+1}^i.
	\end{equation}
	We propagate the refined flow and features along to next level as shown in Figure \ref{fig:main_arch}.

	\subsection{Temporal Consistency Loss with Occlusion-Aware Masking}
	One of the most important challenges in video restoration is to reduce flickering. The flickering causes unnatural temporal  fluctuations in perceived video, thereby creating an annoying artifact. Hence, maintaining temporal consistency between successive frames is the stringent requirement in video restoration. While current SOTA methods rely intrinsically on motion modeling to compensate flickering, this problem is overlooked in the existing literature. 
	
	In this section, a temporal loss is proposed to deliberately eliminate flickering. It is important to note that flickering artifacts are severely observed in regions of the video that are nearly still (i.e. there is no or small motion across frames). For that reason, the proposed loss leverages on optical flow and occlusion-aware mask to align restored frames effectively. This loss computes the difference between reconstructed reference frame $\tilde{I}_t$ and corresponding warped frame  $\tilde{I}^{warp}_{t-1}$ only for regions where there is no pixel-value variation. 
	
	The loss computation is performed in three stages. First, the optical flow $\Delta_{t, t-1}$ is computed between ground truth images $I_t^{HQ}$ and $I_{t-1}^{HQ}$. It is worth noting that the optical flow is computed between clean images since loss function is used only during training. Afterwards, the restored frame $\tilde{I}_{t-1}$ is warped towards the restored reference frame using optical flow. It can be represented as,
	\begin{equation}
		\tilde{I}^{warp}_{t-1} = \text{Warp}(\tilde{I}_{t-1}, \Delta_{t, t-1}).
	\end{equation}
	
	In the second stage, we compute occlusion-aware mask with the ground truth frames. This mask is computed aiming to focus on regions across frames that are not occluded (i.e. regions that are visible in both frames). It is given as,
	\begin{equation}
		m_{t, t-1} = \exp(-\alpha \left\| I_t^{HQ} -  I_{t-1}^{HQ, warp} \right\|_2^2),
	\end{equation}
	where $I_{t-1}^{HQ, warp}=\text{Warp}(I_{t-1}^{HQ}, \Delta_{t, t-1})$. Here, the parameter $\alpha$ should be sufficiently large to allow fast transition from $1$ to $0$ and efficiently penalize the large and non-zero frame differences.  
	
	In the last stage, we compute the distance between $\tilde{I}_t$ and $\tilde{I}^{warp}_{t-1}$. Then we apply mask on the computed distance, thus obtaining loss value for non-occluded regions:
	\begin{equation}
		\mathcal{L}_{t, t-1}^{T} = m_{t, t-1} \odot \left\| \tilde{I}_t - \tilde{I}^{warp}_{t-1} \right\|_1,
	\end{equation} 
	where operator $\odot$ performs element-wise multiplication.
	\subsection{Total Loss}
	The total loss is combination of the temporal consistency loss and well-known $L_1$ distance (absolute error) between reconstructed and ground-truth frame. 
	
	\begin{equation}
		\mathcal{L}_{Total} = \left\| I_t^{HQ} -  \tilde{I}_{t} \right\|_1 + w_1 . \mathcal{L}_{t, t-1}^{T} + w_2 . \mathcal{L}_{t, t+1}^{T},
	\end{equation}
	where $w_1$ and $w_2$ are weighting factors that balance the trade-off between $L_1$ loss and temporal consistency loss. We set $w_1$ and $w_2$ to 0.2 in our experiments. Refer to supplementary material for visual results that demonstrate the significance of this loss function in reducing flickering artifacts.
	
	\section{Experiments}
	In this section, we report a detailed experimental analysis to test the effectiveness of the proposed JFFRA framework on various video restoration tasks. The ablation studies are carried out to signify the contribution of each component of the proposed framework. The performance analysis of the proposed method in comparison with SOTA methods is presented for video denoising, video delburring and video super-resolution tasks. The performance is evaluated in terms of PSNR and SSIM.
	
	\subsection{Experimental Setup}
	For all experiments, the stack of 3 consecutive frames is used as the input. Attention is computed on a window size of 8. The network is trained using the Adam optimizer for 700k iterations.
	The learning rate of the optimizer is set to $1e-4$ which decreases to $1e-6$ using step learning rate scheduler. The network is trained with mini-batches of size 48 and hyper-parameter $\alpha$ is set to 0.2. All experiments are performed on 8 Tesla P40 cards with 24 GB of memory.
	
	\subsection{Video Denoising}
	\paragraph*{4.2.1 Real Denoising Dataset:} The CRVD dataset \cite{yue2020supervised} is captured in raw domain and converted  to sRGB domain using image signal processing module. The dataset contains 6 indoor scenes for training and 5 indoor scenes for testing. For each scene, the dataset has 5 different ISO settings ranging from 1600 to 25600. In \autoref{tab:comp_crvd}, performance of the proposed method on CRVD dataset under different ISO settings is presented. It can be seen that the proposed method achieves best metrics across all ISO settings. The most encouraging finding is that the proposed method beats previous SOTA method by 1.62 dB in PSNR. This improvement in the performance can be attributed to JFFR module for jointly refining the flow and features, thereby efficiently handling temporal information.

	\begin{table*}
		\vspace{-0.2cm}
		\caption{Quantitative analysis of the proposed JFFRA network with SOTA methods for video denoising on the CRVD (RGB) datasets \cite{yue2020supervised}. The best and second best results are highlighted in red and blue colors, respectively.}
		\label{tab:comp_crvd}
		\centering
		\resizebox{1\textwidth}{!}{
			\begin{tabular}{|c|c|c|c|c|c|c|c|c|c|}
				\hline
				Metrics & VBM4D \cite{maggioni2012bm4d}  & ViDeNN \cite{claus2019videnn} & TOFlow \cite{xue2019toflow} & SMD \cite{chen2019smd} & EDVR \cite{wang2019edvr} & DIDN \cite{yu2019didn} & RViDeNet \cite{yue2020supervised} &\bf{Ours} \\
				 & TIP'12 & CVPR'19 & IJCV'19 & ICCV'19 & CVPR'19 & CVPR'19 & CVPR'20 &  \\
				\hline
				PSNR   & 34.16 & 31.48 & 34.81 & 35.87 & 38.97 & 38.83 & \textcolor{blue}{39.95} & \textcolor{red}{41.57} \\ 
				\hline
				SSIM   & 0.922 & 0.826 & 0.921 & 0.957 & 0.972 & 0.974 & \textcolor{blue}{0.979} & \textcolor{red}{0.986} \\ 
				\hline
			\end{tabular}\vspace{-2mm}
		}
	\end{table*}
	
	
	\begin{table*}
		
		\caption{Quantitative analysis with different existing methods for video denoising on the DAVIS \cite{khoreva2019video} and Set-8 \cite{tassano2019dvdnet} datasets. The parameter $\sigma$ indicates the noise level. The best and second best results are highlighted in red and blue colors, respectively.}
		\label{tab:comp_AWGN}
		\centering
		\resizebox{1\textwidth}{!}{
			\begin{tabular}{|c|c|c|c|c|c|c|c|c|c|c|c|c|c|}
				\hline
				Datasets & $\sigma$ & VBM4D \cite{maggioni2012bm4d} & VNLB \cite{arias2018video} & DVDnet \cite{tassano2019dvdnet} & FastDVDnet \cite{tassano2020fastdvdnet} & VNLNet \cite{davy2018vnlnet} & PaCNet \cite{vaksman2021pacnet} & BasicVSR++ \cite{chan2021basicvsr}  & R-VRT \cite{liang2022recurrent} & Tempformer \cite{song2022tempformer} & ShiftNet \cite{li2023simple} & VRT \cite{liang2024vrt} & \textbf{Ours} \\

				 &  & TIP'12 & JMIC'18 & ICIP'19 & CVPR'20 & ICIP'2019 & ICCV'21 & CVPR'19 & NeurIPS'22 & ECCV'22 & CVPR'23  & TIP'24 & - \\
				\hline 
				DAVIS \cite{khoreva2019video} & 10 & 37.58/- & 38.85/- & 38.13/.9657 & 38.71/.9672 & 39.56/.9707 & 39.97/.9713 & 40.13/.9754  & 40.57/- & 40.17/- & \textcolor{blue}{40.91}/-  & {40.82}/\textcolor{blue}{.9776} &  \textcolor{red}{41.13}/\textcolor{red}{.9781} \\
				& 20 & 33.88/- & 35.68/- & 35.70/.9422 & 35.77/.9405 & 36.53/.9464 & 37.10/.9470 & 37.41/.9598 & 38.05/- & 37.36/- & \textcolor{blue}{38.34}/-  & {38.15}/\textcolor{blue}{.9625} & \textcolor{red}{38.54}/\textcolor{red}{.9652} \\
				& 30 & 31.65/- & 33.73/- & 34.08/.9188 & 34.04/.9167 & -/- & 35.07/.9211 & 35.74/.9456 & 36.57/- &  35.66/- & \textcolor{blue}{36.83}/-  & {36.52}/\textcolor{blue}{.9483} & \textcolor{red}{37.04}/\textcolor{red}{.9569} \\
				& 40 & 30.05/- & 32.32/- & 32.86/.8962 & 32.82/.8949 & 33.32/.8996 & 33.57/.8969 & 34.49/.9319  & 35.47/- & 34.42/- & \textcolor{red}{35.71}/- & {35.32}/\textcolor{blue}{.9345} & \textcolor{blue}{35.67}/\textcolor{red}{.9362} \\
				& 50 & 28.80/- & 31.13/- & 31.85/.8745 & 31.86/.8747 & -/- & 32.39/.8743 & 33.45/.9179  & 34.57/- & 33.44/- & \textcolor{red}{34.82}/- & {34.36}/\textcolor{blue}{.9211} &  \textcolor{blue}{34.65}/\textcolor{red}{.9321} \\
				\hline
				Set-8 \cite{tassano2019dvdnet} & 10 & 36.05/- & 37.26/- & 36.08/.9510 & 36.44/.9540 & 37.28/.9606 & 37.06/.9590 & 36.83/.9574  & 37.53/- & 37.15/- & 37.48/- & \textcolor{blue}{37.88}/\textcolor{blue}{.9630} & \textcolor{red}{38.17}/\textcolor{red}{.9642} \\
				& 20 & 32.18/- & 33.72/- & 33.49/.9182 & 33.43/.9196 & 34.02/.9273 & 33.94/.9247 & 34.15/.9319  & 34.83/- & 34.74/- & 34.85/-& \textcolor{blue}{35.02}/\textcolor{blue}{.9373} &  \textcolor{red}{35.22}/\textcolor{red}{.9386} \\
				& 30 & 30.00/- & 31.74/- & 31.68/.8862 & 31.68/.8889 & -/- & 32.05/.8921 & 32.57/.9095  & 33.30/- & 33.20/- & 33.29/-  & \textcolor{red}{33.35}/\textcolor{red}{.9141} & \textcolor{blue}{33.30/ .9118} \\
				& 40 & 28.48/- & 30.39/- & 30.46/.8564 & 30.46/.8608 & 30.72/.8622 & 30.70/.8623 & 30.42/.8589  & \textcolor{red}{32.21}/- & 32.06/- & \textcolor{blue}{32.18}/- &  {32.15}/\textcolor{red}{.8928} &  {32.06}/\textcolor{blue}{.8842} \\
				& 50 & 27.33/- & 29.24/- & 29.53/.8289 & 29.53/.8351 & -/- & 29.66/.8349 & 30.49/.8690 & \textcolor{red}{31.33}/- & 31.16/- & {31.33}/-  & {31.22}/\textcolor{red}{.8733} & \textcolor{blue}{31.24/.8726}  \\
				\hline

			\end{tabular}
		}
	\end{table*}
\begin{figure*}[!htbp]
	\captionsetup{font=small}
	
	\centering
	\scriptsize
	
	\renewcommand{\h}{0.105}
	\renewcommand{\wa}{0.12}
	\newcommand{\wb}{0.16}
	\renewcommand{\g}{-0.7mm}
	\renewcommand{\tabcolsep}{1.8pt}
	\renewcommand{\arraystretch}{1}
	\resizebox{1.00\linewidth}{!} {
		\begin{tabular}{cc}
			\renewcommand{\name}{davis/}
			\renewcommand{\h}{0.15}
			\renewcommand{\w}{0.2}
			\begin{tabular}{cc}
				\begin{adjustbox}{valign=t}
					\begin{tabular}{c}%
						\includegraphics[trim={25 0 0 0 },clip, width=0.354\textwidth]{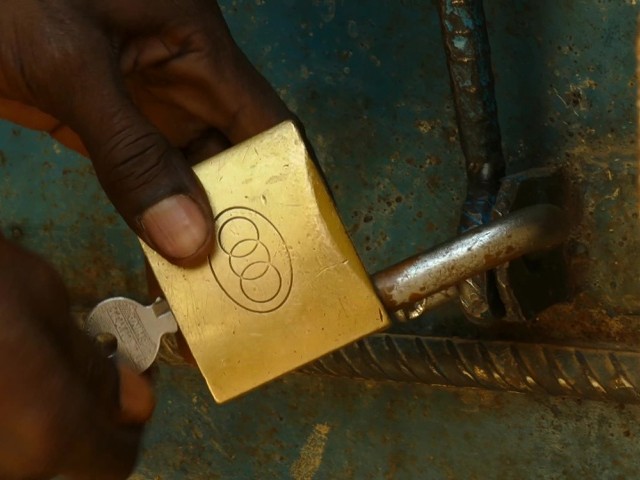}
						\\
						Frame 32, Lock scene, DAVIS Dataset \cite{khoreva2019video}
					\end{tabular}
				\end{adjustbox}
				\begin{adjustbox}{valign=t}
					\begin{tabular}{cccccc}
						\includegraphics[height=\h \textwidth, width=\w \textwidth]{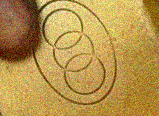} \hspace{\g} &
						\includegraphics[height=\h \textwidth, width=\w \textwidth]{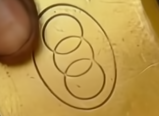} \hspace{\g} &
						\includegraphics[height=\h \textwidth, width=\w \textwidth]{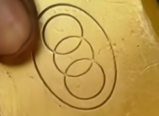} 
						\\
						LQ ($\sigma = 20$) &
						VRT \cite{liang2024vrt}  &
						RVRT \cite{liang2022recurrent} 
						\\
						\vspace{-1.5mm}
						\\
						\includegraphics[height=\h \textwidth, width=\w \textwidth]{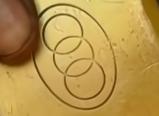} \hspace{\g} &
						\includegraphics[height=\h \textwidth, width=\w \textwidth]{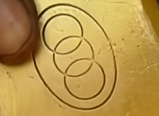} \hspace{\g} &
						\includegraphics[height=\h \textwidth, width=\w \textwidth]{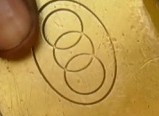} 
						\\ 
						ShiftNet~\cite{song2022tempformer} \hspace{\g} &
						JFFRA (ours) &
						GT
						\\
					\end{tabular}
				\end{adjustbox}
			\end{tabular}
			
		\end{tabular}
	}
	\vspace{-2mm}
	\caption{Visual comparison on video denoising task. The JFFRA effectively remove noise while preserving structure and sharpness.} %
	\vspace{-2mm}
	\label{fig:denoise_img}
\end{figure*}
	
	\paragraph*{4.2.2 Synthetic Video Denoising :} DAVIS \cite{khoreva2019video} and Set-8 \cite{tassano2019dvdnet} are the standard synthetic video denoising datasets. While DAVIS consists of 90 training and 30 testing sequences, Set-8 have 8 testing video sequences. Following the synthetic pipeline setting as VRT \cite{liang2024vrt}, noisy sequences are synthesized by adding AWGN with noise level \(\sigma \in \) [0,50] on the DAVIS training set.
	
	The evaluation is performed on Davis and Set-8 test data with noise levels $\sigma=\{10,20,30,40,50\}$. It can be observed from \autoref{tab:comp_AWGN}  that \mypapertitle outperforms existing methods on DAVIS dataset while giving competitive performance on Set-8 dataset.
	Additionally, it can be visualized from  \figurename~\ref{fig:denoise_img} that the proposed method denoises video frames while maintaining the texture and high frequency regions compared to other methods.
	
	\subsection{Video Super Resolution}
\begin{figure*}[!htbp]
	\captionsetup{font=small}
	
	\centering
	\scriptsize
	
	\renewcommand{\h}{0.105}
	\renewcommand{\wa}{0.12}
	\newcommand{\wb}{0.16}
	\renewcommand{\g}{-0.7mm}
	\renewcommand{\tabcolsep}{1.8pt}
	\renewcommand{\arraystretch}{1}
	\resizebox{1.00\linewidth}{!} {
		\begin{tabular}{cc}
			
			\renewcommand{\name}{figures/sr/00000024_}
			\renewcommand{\h}{0.12}
			\renewcommand{\w}{0.2}
			\begin{tabular}{cc}
				\begin{adjustbox}{valign=t}
					\begin{tabular}{c}%
						\includegraphics[trim={336 0 0 0 },clip, width=0.354\textwidth]{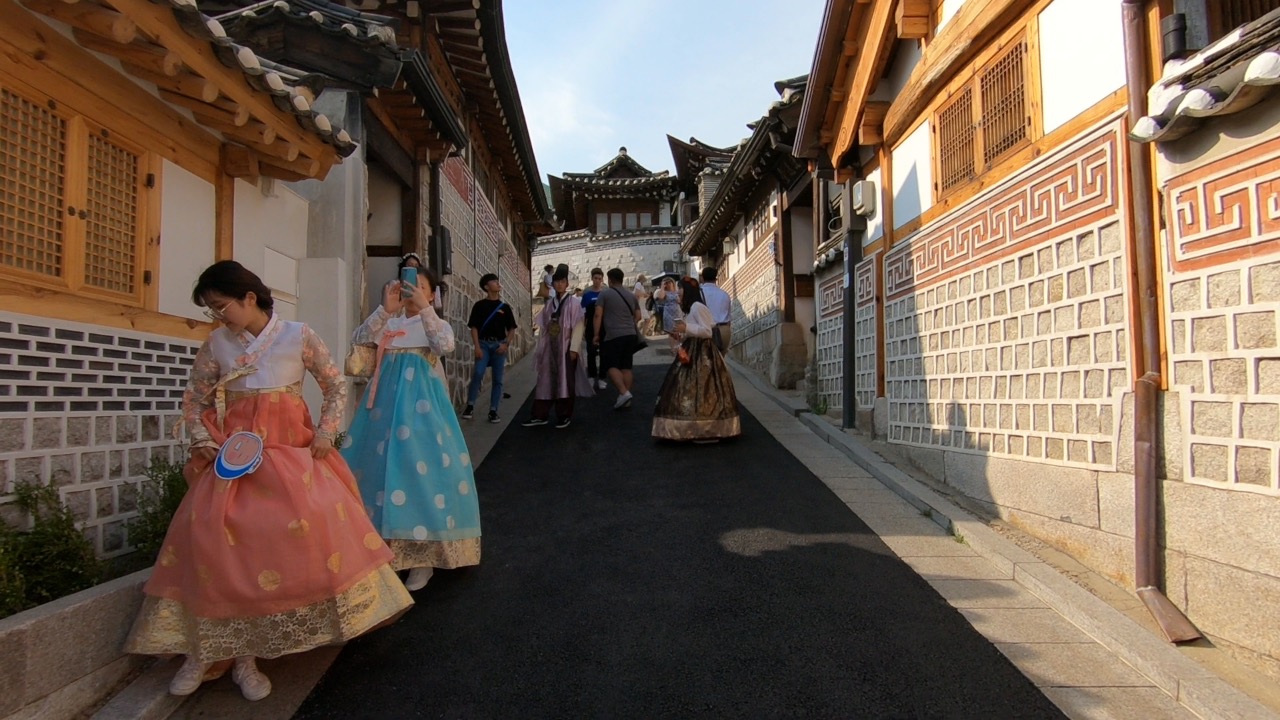}
						\\
						Frame 024, Clip 011, REDS  dataset \cite{nah2018deepmultiscaleconvolutionalneural}
					\end{tabular}
				\end{adjustbox}
				\begin{adjustbox}{valign=t}
					\begin{tabular}{cccccc}
						\includegraphics[height=\h \textwidth, width=\w \textwidth]{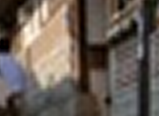} \hspace{\g} &
						\includegraphics[height=\h \textwidth, width=\w \textwidth]{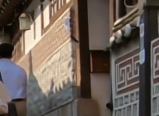} \hspace{\g} &
						\includegraphics[height=\h \textwidth, width=\w \textwidth]{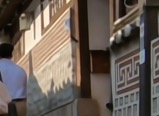} 
						\\
						LQ ($\times 4$) &
						BasicVSR++ \cite{chan2021basicvsrpp} &
						VRT \cite{liang2024vrt} 
						\\
						\vspace{-1.5mm}
						\\
						\includegraphics[height=\h \textwidth, width=\w \textwidth]{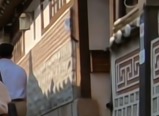} \hspace{\g} &
						\includegraphics[height=\h \textwidth, width=\w \textwidth]{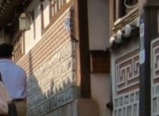} \hspace{\g} &
						\includegraphics[height=\h \textwidth, width=\w \textwidth]{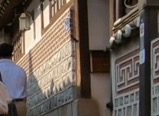} 
						\\ 
						RVRT \cite{liang2022recurrent} \hspace{\g} &
						JFFRA (ours) &
						GT
						\\
					\end{tabular}
				\end{adjustbox}
			\end{tabular}
			
		\end{tabular}
	}
	\vspace{-2mm}
	\caption{Visual comparison on video super-resolution ($\times 4$) task. The proposed JFFRA recovers wall structure and fine patterns efficiently.}  %
	\vspace{-2mm}
	\label{fig:vsr_img}
\end{figure*}

	\vspace{-1.5mm}
	The comparative analysis of various methods on video super-resolution $(\times 4)$ task is detailed in \autoref{tab:simple_comparison}. The network is trained with both bi-cubic (BI) and blur-down sampling (BD) degradations on standard datasets such as REDS (RGB) \cite{nah2018deepmultiscaleconvolutionalneural} and Vimeo-90K  \cite{xue2019video}. The performance is evaluated our Vimeo-90k, REDS, Vid4 \cite{arias2018video}, and UDM10 \cite{yi2019progressive} dataset. It can be noticed that the proposed method achieves the best performance on both BI and BD degradations. While JFFRA outperforms SOTA VRT \cite{liang2024vrt} by 0.26 $\mathtt{\sim}$ 0.6 dB on REDS and  Vid4 dataset, it surpasses VRT by 1 $\mathtt{\sim}$ 1.3 dB on  Vimeo-90K and UDM10 dataset.This performance improvement signifies that the proposed offset flow learning and residual feature refinement guides the network to converge to a better optimal minimum. The qualitative results are displayed in \figurename~\ref{fig:vsr_img} for visual analysis. It can be noticed that the proposed method is able 	to retain texture more precisely and robust to detail loss as compared to other methods.
	
	\begin{table*}
		\vspace{-4mm}
		\caption{Quantitative analysis of different methods in terms of PSNR/SSIM for video super-resolution (×4) using BI and BD degradation settings on various datasets. The best and second best results are highlighted in red and blue colors, respectively.}
		\label{tab:simple_comparison}
		\centering
		\resizebox{1\textwidth}{!}{ 
			\begin{tabular}{|l|c|c|c|c|c|c|c|c|}
				\hline
				\textbf{Method} &
				\textbf{Publication} & \textbf{Training Frames (REDS/Vimeo-90K)} & \multicolumn{3}{c|}{\textbf{BI Degradation}} & \multicolumn{3}{c|}{\textbf{BD Degradation}} \\ \hline
				&  & & \textbf{REDS4 (RGB) \cite{nah2018deepmultiscaleconvolutionalneural}} & \textbf{Vimeo-90K-T (Y)} ~\cite{xue2019video} & \textbf{Vid4 (Y)}  ~\cite{arias2018video} & \textbf{UDM10 (Y)} ~\cite{yi2019progressive} & \textbf{Vimeo-90K-T (Y)} ~\cite{xue2019video} & \textbf{Vid4 (Y)} ~\cite{arias2018video} \\ \hline
				Bicubic & - & - & 26.14/0.7292 & 31.32/0.8684 & 23.78/0.6347 & 28.47/0.8253 & 31.30/0.8687 & 21.80/0.5246 \\ \hline
				SwinIR ~\cite{liang2021swinir} & ICCV'21  & - & 29.05/0.8269 & 35.67/0.9287 & 25.68/0.7491 & 35.42/0.9380 & 34.12/0.9167 & 25.25/0.7262 \\ \hline
				TOFlow ~\cite{xue2019toflow} & IJCV'19 & 5/7 & 27.98/0.7990 & 33.08/0.9054 & 25.89/0.7651 & 36.26/0.9438 & 34.62/0.9212 & 25.85/0.7659 \\ \hline
				DUF ~\cite{jo2018deep} & CVPR'18 & 7/7 & 28.63/0.8251 & - & 27.33/0.8319 & 38.48/0.9605 & 36.87/0.9447 & 27.38/0.8329 \\ \hline
				PFNL ~\cite{yi2019progressive} & ICCV'19 & 7/7 & 29.63/0.8502 & 36.14/0.9363 & 26.73/0.8029 & 38.74/0.9627 & - & 27.16/0.8355 \\ \hline
				RBPN ~\cite{haris2019recurrent} & CVPR'19 & 7/7 & 30.09/0.8590 & 37.07/0.9435 & 27.12/0.8180 & 38.66/0.9596 & 37.20/0.9458 & 27.17/0.8205 \\ \hline
				EDVR ~\cite{wang2019edvr} & CVPR'19 & 5/7 & 31.09/0.8800 & 37.61/0.9489 & 27.35/0.8264 & 39.89/0.9686 & 37.81/0.9523 & 27.85/0.8503 \\ \hline
				BasicVSR ~\cite{chan2021basicvsr} & CVPR'21 & 15/14 & 31.42/0.8909 & 37.18/0.9450 & 27.24/0.8251 & 39.96/0.9694 & 37.53/0.9498 & 27.96/0.8553 \\ \hline
				IconVSR ~\cite{chan2021basicvsr} & CVPR'21 & 15/14 & 31.67/0.8948 & 37.47/0.9476 & 27.39/0.8279 & 40.03/0.9694 &
				37.84/0.9524 & 28.04/0.8570 \\ \hline
				BasicVSR++ ~\cite{chan2021basicvsrpp} & CVPR'22 & 30/14 & {32.39/0.9069} & 37.79/0.9500 & {27.79/0.8400} & 40.72/0.9722 & 38.21/0.9550 & 29.04/0.8753 \\ \hline			
				RVRT ~\cite{liang2022recurrent} & NeurIPS'22 & 16/7 &
				 \textcolor{red}{32.75/0.9113} &
				 {38.15/0.9527} &
				 \textcolor{blue}{27.99/0.8462} &
				 {40.90/0.9729} &
				 {38.59/0.9576} & 
				 \textcolor{blue}{29.54/0.8810} \\ \hline
				VRT ~\cite{liang2024vrt} & TIP'24 & 16/7 & 32.19/0.9006 & \textcolor{blue}{38.20/0.9530} & {27.9/0.8425} & \textcolor{blue}{41.05/0.9737} & \textcolor{blue}{38.72/0.9584} & {29.42/0.8795} \\ \hline 
				JFFRA (ours) & - & 3/3 & \textcolor{blue}{32.45/0.9136} & \textcolor{red}{39.32/0.9561} & \textcolor{red}{28.54/0.8725} & \textcolor{red}{42.15/0.9848} & \textcolor{red}{39.62/0.9786} & \textcolor{red}{30.52/0.8896} \\ \hline
			\end{tabular}
		}\vspace{-2mm}
	\end{table*}
	
	
	\begin{table*}
		\caption{Quantitative analysis of various methods in terms of PSNR/SSIM for video deblurring on  DVD \cite{su2017deep}. The best and second best results are highlighted in red and blue colors, respectively.}
		\label{tab:comp_deblur}
		\centering
		\resizebox{1\textwidth}{!}{
			\begin{tabular}{|c|c|c|c|c|c|c|c|c|c|c|c|c|c|c|c|}
				\hline
				Metric & DBN \cite{su2017deep} & DBLRNet \cite{zhang2018adversarial} 
				& STFAN \cite{zhou2019spatio} & STTN \cite{kim2018spatio} & SFE \cite{xiang2020deep} 
				& EDVR \cite{wang2019edvr} & TSP \cite{pan2020cascaded} & PVDNet \cite{son2021recurrent} 
				& GSTA \cite{suin2021gated} & ARVo \cite{li2021arvo}  & R-VRT \cite{liang2022recurrent} & VRT \cite{liang2024vrt} & ShiftNet \cite{li2023simple} & BSSTNet \cite{zhang2024blur} &\bf{Ours} \\
				& CVPR'17 & TIP'18 & ICCV'19 & ECCV'18 & TIP'20 & CVPR'19 & CVPR'20 & TOG'21 & CVPR'21 & CVPR'21 & NuerIPS'22  & TIP'24 & CVPR'23 & CVPR'24 &  \\
				\hline
				PSNR & 30.01 & 30.08 & 31.24 & 31.61 & 31.71 & 31.82 & 32.13 & 32.31 & 32.53 & 32.80 & 34.30 & {34.27} & 34.69 & \textcolor{blue}{34.95} & \textcolor{red}{35.15} \\ 
				\hline
				SSIM & 0.8877 & 0.8845 & 0.9340 & 0.9160 & 0.9160 & 0.9160 & 0.9268 & 0.9260 & 0.9468 & 0.9352  & 0.9655 & {0.9651} & 0.969 & \textcolor{blue}{0.97} & \textcolor{red}{0.976} \\ 
				\hline
			\end{tabular}
		}
	\end{table*}

	\begin{figure*}[!htbp]
	\captionsetup{font=small}
	
	\centering
	\scriptsize
	
	\renewcommand{\h}{0.105}
	\renewcommand{\wa}{0.12}
	\newcommand{\wb}{0.16}
	\renewcommand{\g}{-0.7mm}
	\renewcommand{\tabcolsep}{1.8pt}
	\renewcommand{\arraystretch}{1}
	\resizebox{1.00\linewidth}{!} {
		\begin{tabular}{cc}
			
			\renewcommand{\name}{DVD/}
			\renewcommand{\h}{0.14}
			\renewcommand{\w}{0.2}
			\begin{tabular}{cc}
				\begin{adjustbox}{valign=t}
					\begin{tabular}{c}%
						\includegraphics[trim={336 0 0 0 },clip, width=0.354\textwidth]{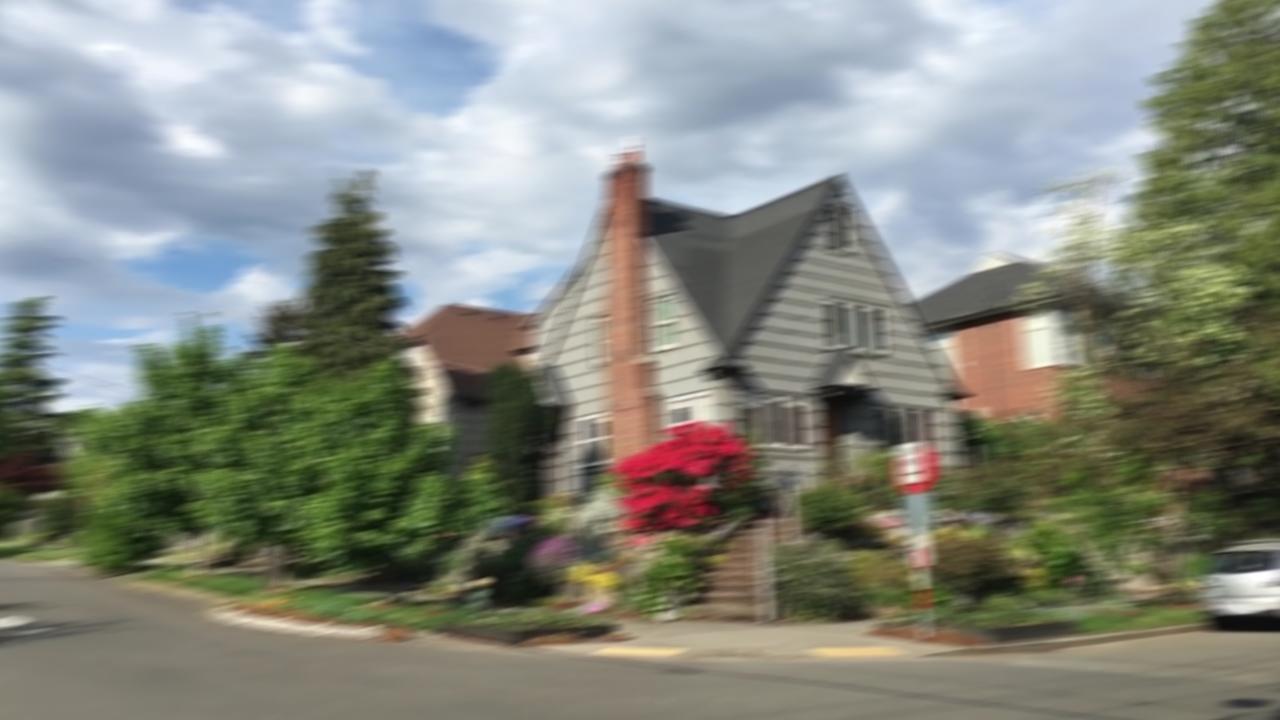}
						\\
						Frame 0034, Clip IMG 0030, DVD dataset \cite{su2017deep}
					\end{tabular}
				\end{adjustbox}
				\begin{adjustbox}{valign=t}
					\begin{tabular}{cccccc}
						\includegraphics[height=\h \textwidth, width=\w \textwidth]{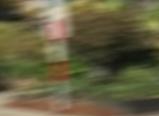} \hspace{\g} &
						\includegraphics[height=\h \textwidth, width=\w \textwidth]{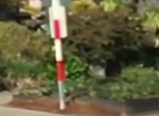} \hspace{\g} &
						\includegraphics[height=\h \textwidth, width=\w \textwidth]{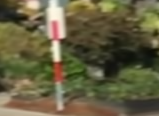} 
						\\
						LQ  &
						RVRT \cite{liang2022recurrent} &
						ShiftNet \cite{li2023simple} 
						\\ 
						\vspace{-2mm}
						\\
						\includegraphics[height=\h \textwidth, width=\w \textwidth]{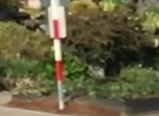} \hspace{\g} &
						\includegraphics[height=\h \textwidth, width=\w \textwidth]{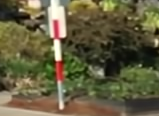} \hspace{\g} &
						\includegraphics[height=\h \textwidth, width=\w \textwidth]{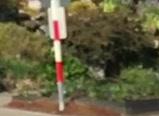} 
						\\ 
						BSSTNet \cite{zhang2024blur} \hspace{\g} &
						{JFFRA} (ours) &
						GT
						\\
					\end{tabular}
				\end{adjustbox}
			\end{tabular}
			
		\end{tabular}
	}
	\vspace{-2mm}
	\caption{Visual comparison on video deblurring task. The restored image using JFFRA is more sharper and closer to ground truth.} %
	\vspace{-2mm}
	\label{fig:deblur_img}
\end{figure*}

	\subsection{Video Deblurring}
	\vspace{-1.5mm}
	The usefulness of the proposed method in practical applications is demonstrated by performing experiments on video deblurring task. The experiments are conducted on the DVD \cite{su2017deep} dataset. It is clear from the results shown in \autoref{tab:comp_deblur} that the proposed method yields promising results and attains SOTA performance. The proposed method surpasses ShiftNet \cite{li2023simple} and  BSSTNet \cite{zhang2024blur} by 0.46 dB and 0.2 dB respectively. \figurename~\ref{fig:deblur_img} shows the visual comparison between various methods. It can be visualized that the proposed method effectively removes the blur while restoring the fine details. This observation is consistent with the improved quantitative performance.
	
	\subsection{Computational Cost}
\begin{figure}[H]
		\centering
		\includegraphics[width=0.35\textwidth,height=0.20\textwidth]{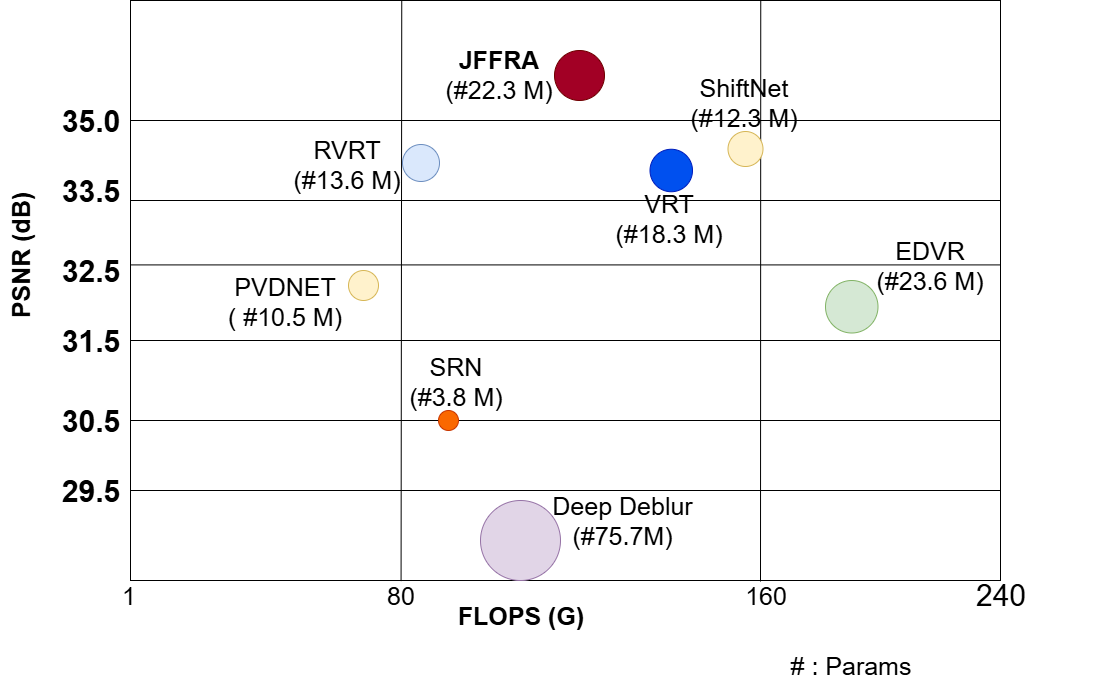}
		\caption{Complexity Analysis of JFFRA with other state-of-the-art networks on Video Deblurring Task.}
		\label{fig:Complexity}
		\vspace{-2mm}
	\end{figure}

	\vspace{-2mm}
	\figurename~\ref{fig:Complexity} shows that our method achieves the optimal
	trade-off between computational cost and restoration performance. The complexity of JFFRA is much lower than VRT \cite{liang2024vrt}, ShiftNet \cite{li2023simple} while achieving much better performance. Refer supplementary for detailed analysis.  
	
	\section{Ablation Studies}
	We conducted several experiments to demonstrate the effectiveness of the proposed JFFR block and temporal loss function.

	{\textbf{Impact of JFFR Module :}}
	We replace the proposed block with four variations: 1) Mutual self-attention substituting without parallel warping similar to \cite{liang2024vrt}, 2) attention with learnable offsets, 3) attention without refined flow, and 4) attention with iterative refined flow. The quantitative results on the DVD deblur dataset as detailed in \autoref{tab:comp_deformable}, reveal that joint refinement of flow and feature in iterative manner complement each other, thereby significantly improving the performance.

	\begin{table}[H]
		\caption{Ablation results on proposed JFFR Module.}
		\label{tab:comp_deformable}
		\centering
		\resizebox{0.45\textwidth}{!}{
			\begin{tabular}{|c|c|c|}
				\hline
				Methods & PSNR & SSIM \\
				\hline
				Temporal mutual self attention & 28.5 & 0.87 \\ 
				\hline
				Attention with learnable offset & 28.72 & 0.884 \\ 
				\hline
				Attention without refined flow & 29.45 &0.897 \\ 
				\hline
				JFFR  & 35.15 & 0.976 \\ 
				\hline					
			\end{tabular}
		}
	\end{table}
	
\begin{figure}[H]
		\centering
		\includegraphics[width=0.43\textwidth,height=0.45\textwidth]{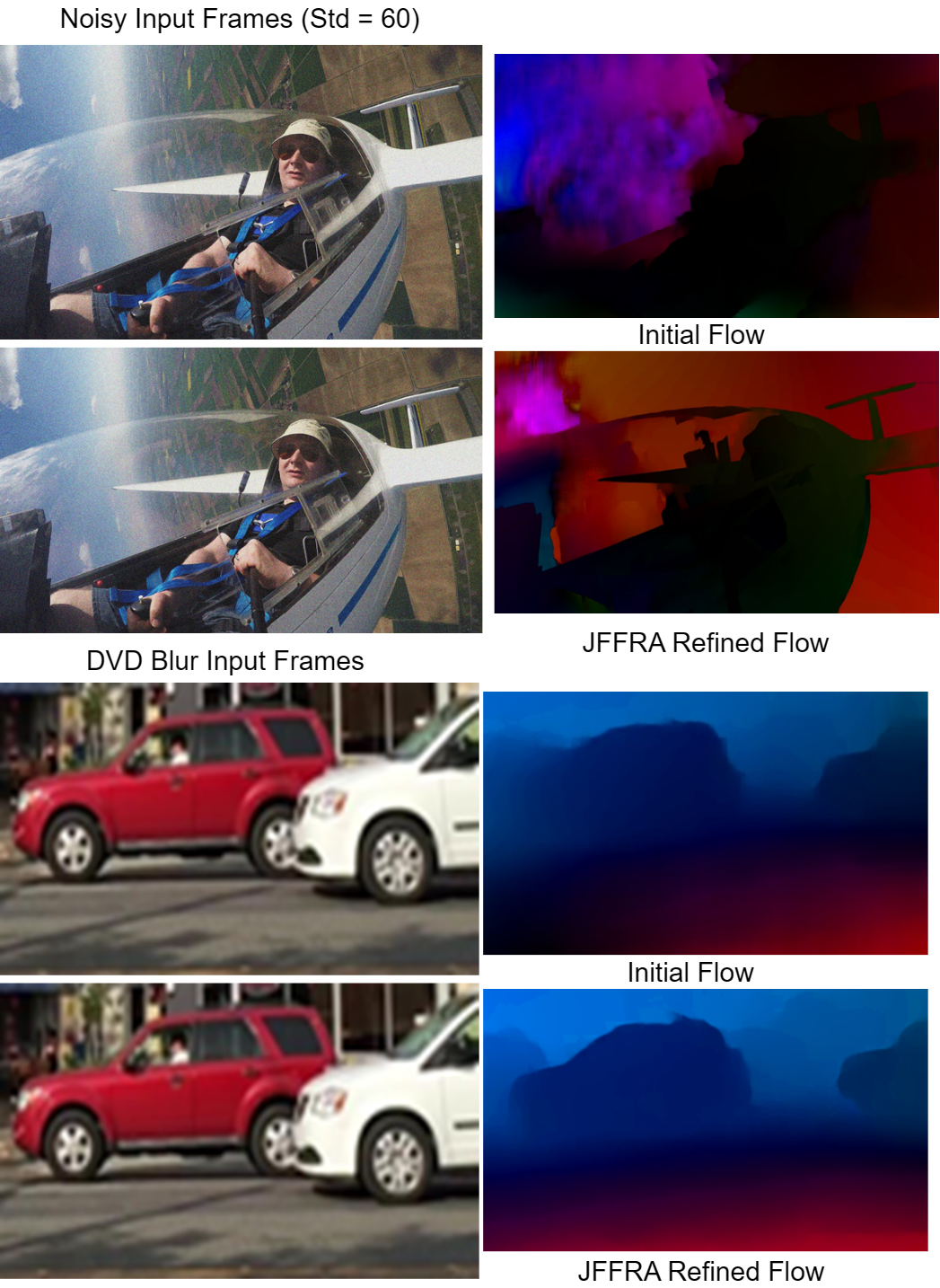}
		\caption{Impact of flow refinement in retrieving the information from low quality frames. Frames used are Davis-Test \cite{khoreva2019video} (aerobatics frame 13, 14) and DVD \cite{su2017deep} (IMG0037, frame 13,14)}
		\label{fig:impactoflfow}
		\vspace{-5mm}
	\end{figure}

	{\textbf{Impact of Flow Refinement Module :}}
	Many researchers have explored the use of optical flow for video restoration tasks. In all the previous works, flow is computed either on degraded input or initial recovered frames, without continuous refinement.To demonstrate the effectiveness of iterative flow refinement, we initially trained our network without flow refinement module. It can be clearly seen from \autoref{tab:comp_deformable} that lack of flow refinement causes heavy losses in the restoration.This poor recovery performance can be attributed to fundamental flaws in the initial flow. We further showcase in  \figurename~\ref{fig:impactoflfow} the impact of flow refinement. It can be deduced from \autoref{tab:comp_deformable} and \figurename~\ref{fig:impactoflfow} that joint refinement of frame features and flow together with their complementary enhancement fuels persistent improvement.

	{\textbf{Impact of Temporal Loss Function :}}
	The importance of temporal loss function in reducing flickering artifacts is summarized in \autoref{tab:opw}. The temporal consistency between successive frames is measured using the optical flow based warping metric (OPW) following FMNet \cite{xian2018monocular} and DPT \cite{wang2023neural}.	The efficiency of the proposed temporal loss function in reducing inconsistency between successive frames can be concluded  from \autoref{tab:opw}. Refer to supplementary material for more detailed and visual analysis.

	\begin{table}[H]
		\caption{Ablation results to show usefulness of the proposed temporal loss function.}
		\label{tab:opw}
		\centering
		\resizebox{0.35\textwidth}{!}{
			\begin{tabular}{|c|c|}
				\hline
				Methods & OPW\\
				\hline
				JFFRA without temporal loss &  0.372\\ 
				\hline
				JFFRA with temporal loss & 0.354 \\ 
				\hline
			\end{tabular}
		}
	\end{table}
	
	\section{Conclusions}
	In this paper, we propose a novel joint flow and feature refinement using attention framework for video restoration. Unlike previous end-to-end learning approaches, JFFRA utilizes an iterative refinement process that builds on collaborative synergy between flow and restoration. At each stage, the flow refinement module refines the current flow field based on existing enhanced features information, while the feature refinement block uses the refined flow to further enhance feature alignment and information fusion. Through this mutual collaboration, JFFRA effectively addresses the limitations of traditional methods. Comprehensive experiments on various restoration tasks and benchmark datasets validate the versatility and superiority of JFFRA, establishing its position as a leading solution in the field. With up to 1.62 dB gain in PSNR compared to state-of-the-art methods, JFFRA sets a new standard for video restoration algorithms.
	\FloatBarrier

\newcommand{\myformattingnote}{}



{
    \small
    \bibliographystyle{ieeenat_fullname}
    \bibliography{main}

\begin{thebibliography}{62}
\providecommand{\natexlab}[1]{#1}
\providecommand{\url}[1]{\texttt{#1}}
\expandafter\ifx\csname urlstyle\endcsname\relax
  \providecommand{\doi}[1]{doi: #1}\else
  \providecommand{\doi}{doi: \begingroup \urlstyle{rm}\Url}\fi

\bibitem[Argaw et~al.(2021)Argaw, Kim, Rameau, Zhang, and
  Kweon]{argaw2021restoration}
Dawit~Mureja Argaw, Junsik Kim, Francois Rameau, Chaoning Zhang, and In~So
  Kweon.
\newblock Restoration of video frames from a single blurred image with motion
  understanding.
\newblock In \emph{Proceedings of the IEEE/CVF Conference on Computer Vision
  and Pattern Recognition}, pages 701--710, 2021.

\bibitem[Arias and Morel(2018)]{arias2018video}
Pablo Arias and Jean-Michel Morel.
\newblock Video denoising via empirical bayesian estimation of space-time
  patches.
\newblock \emph{Journal of Mathematical Imaging and Vision}, 60:\penalty0
  70--93, 2018.

\bibitem[Buades et~al.(2016)Buades, Lisani, and Miladinović]{7448946}
Antoni Buades, Jose-Luis Lisani, and Marko Miladinović.
\newblock Patch-based video denoising with optical flow estimation.
\newblock \emph{IEEE Transactions on Image Processing}, 25\penalty0
  (6):\penalty0 2573--2586, 2016.

\bibitem[Caballero et~al.(2017)Caballero, Ledig, Aitken, Acosta, Totz, Wang,
  and Shi]{caballero2017real}
Jose Caballero, Christian Ledig, Andrew Aitken, Alejandro Acosta, Johannes
  Totz, Zehan Wang, and Wenzhe Shi.
\newblock Real-time video super-resolution with spatio-temporal networks and
  motion compensation.
\newblock In \emph{Proceedings of the IEEE conference on computer vision and
  pattern recognition}, pages 4778--4787, 2017.

\bibitem[Cao et~al.(2021)Cao, Li, Zhang, and Van~Gool]{cao2021vsrt}
Jiezhang Cao, Yawei Li, Kai Zhang, and Luc Van~Gool.
\newblock Video super-resolution transformer.
\newblock \emph{arXiv}, 2021.

\bibitem[Cao et~al.(2022)Cao, Liang, Zhang, Li, Zhang, Wang, and
  Van~Gool]{cao2022datsr}
Jiezhang Cao, Jingyun Liang, Kai Zhang, Yawei Li, Yulun Zhang, Wenguan Wang,
  and Luc Van~Gool.
\newblock Reference-based image super-resolution with deformable attention
  transformer.
\newblock In \emph{European conference on computer vision}, 2022.

\bibitem[Chan et~al.(2021)Chan, Wang, Yu, Dong, and Loy]{chan2021basicvsr}
Kelvin~CK Chan, Xintao Wang, Ke Yu, Chao Dong, and Chen~Change Loy.
\newblock Basicvsr: The search for essential components in video
  super-resolution and beyond.
\newblock In \emph{Proceedings of the IEEE/CVF conference on computer vision
  and pattern recognition}, pages 4947--4956, 2021.

\bibitem[Chan et~al.(2022)Chan, Zhou, Xu, and Loy]{chan2021basicvsrpp}
Kelvin~CK Chan, Shangchen Zhou, Xiangyu Xu, and Chen~Change Loy.
\newblock Basicvsr++: Improving video super-resolution with enhanced
  propagation and alignment.
\newblock In \emph{IEEE Conference on Computer Vision and Pattern Recognition},
  2022.

\bibitem[Chen et~al.(2019)Chen, Chen, Do, and Koltun]{chen2019smd}
Chen Chen, Qifeng Chen, Minh~N Do, and Vladlen Koltun.
\newblock Seeing motion in the dark.
\newblock In \emph{IEEE International Conference on Computer Vision}, pages
  3185--3194, 2019.

\bibitem[Claus and van Gemert(2019)]{claus2019videnn}
Michele Claus and Jan van Gemert.
\newblock Videnn: Deep blind video denoising.
\newblock In \emph{IEEE Conference on Computer Vision and Pattern Recognition
  Workshops}, 2019.

\bibitem[Davy et~al.(2018)Davy, Ehret, Morel, Arias, and
  Facciolo]{davy2018vnlnet}
Axel Davy, Thibaud Ehret, Jean-Michel Morel, Pablo Arias, and Gabriele
  Facciolo.
\newblock Non-local video denoising by cnn.
\newblock \emph{arXiv preprint arXiv:1811.12758}, 2018.

\bibitem[Fischer et~al.(2015)Fischer, Dosovitskiy, Ilg, H{\"a}usser,
  Haz{\i}rba{\c{s}}, Golkov, Van~der Smagt, Cremers, and
  Brox]{fischer2015flownet}
Philipp Fischer, Alexey Dosovitskiy, Eddy Ilg, Philip H{\"a}usser, Caner
  Haz{\i}rba{\c{s}}, Vladimir Golkov, Patrick Van~der Smagt, Daniel Cremers,
  and Thomas Brox.
\newblock Flownet: Learning optical flow with convolutional networks.
\newblock \emph{arXiv preprint arXiv:1504.06852}, 2015.

\bibitem[Haris et~al.(2019)Haris, Shakhnarovich, and Ukita]{haris2019recurrent}
Muhammad Haris, Gregory Shakhnarovich, and Norimichi Ukita.
\newblock Recurrent back-projection network for video super-resolution.
\newblock In \emph{Proceedings of the IEEE/CVF conference on computer vision
  and pattern recognition}, pages 3897--3906, 2019.

\bibitem[Huang et~al.(2015)Huang, Wang, and Wang]{huang2015bidirectional}
Yan Huang, Wei Wang, and Liang Wang.
\newblock Bidirectional recurrent convolutional networks for multi-frame
  super-resolution.
\newblock \emph{Advances in neural information processing systems}, 28, 2015.

\bibitem[Jo et~al.(2018)Jo, Oh, Kang, and Kim]{jo2018deep}
Younghyun Jo, Seoung~Wug Oh, Jaeyeon Kang, and Seon~Joo Kim.
\newblock Deep video super-resolution network using dynamic upsampling filters
  without explicit motion compensation.
\newblock In \emph{Proceedings of the IEEE conference on computer vision and
  pattern recognition}, pages 3224--3232, 2018.

\bibitem[Khoreva et~al.(2019)Khoreva, Rohrbach, and Schiele]{khoreva2019video}
Anna Khoreva, Anna Rohrbach, and Bernt Schiele.
\newblock Video object segmentation with language referring expressions.
\newblock In \emph{Computer Vision--ACCV 2018: 14th Asian Conference on
  Computer Vision, Perth, Australia, December 2--6, 2018, Revised Selected
  Papers, Part IV 14}, pages 123--141. Springer, 2019.

\bibitem[Kim et~al.(2024)Kim, Jeong, Cho, Jeong, and Yoon]{kim2024towards}
Taewoo Kim, Jaeseok Jeong, Hoonhee Cho, Yuhwan Jeong, and Kuk-Jin Yoon.
\newblock Towards real-world event-guided low-light video enhancement and
  deblurring.
\newblock \emph{arXiv preprint arXiv:2408.14916}, 2024.

\bibitem[Kim et~al.(2018)Kim, Sajjadi, Hirsch, and Scholkopf]{kim2018spatio}
Tae~Hyun Kim, Mehdi~SM Sajjadi, Michael Hirsch, and Bernhard Scholkopf.
\newblock Spatio-temporal transformer network for video restoration.
\newblock In \emph{Proceedings of the European conference on computer vision
  (ECCV)}, pages 106--122, 2018.

\bibitem[Li et~al.(2021)Li, Xu, Zhang, Yu, Zhong, Ren, Suominen, and
  Li]{li2021arvo}
Dongxu Li, Chenchen Xu, Kaihao Zhang, Xin Yu, Yiran Zhong, Wenqi Ren, Hanna
  Suominen, and Hongdong Li.
\newblock Arvo: Learning all-range volumetric correspondence for video
  deblurring.
\newblock In \emph{Proceedings of the IEEE/CVF Conference on Computer Vision
  and Pattern Recognition}, pages 7721--7731, 2021.

\bibitem[Li et~al.(2023)Li, Shi, Zhang, Cheung, See, Wang, Qin, and
  Li]{li2023simple}
Dasong Li, Xiaoyu Shi, Yi Zhang, Ka~Chun Cheung, Simon See, Xiaogang Wang,
  Hongwei Qin, and Hongsheng Li.
\newblock A simple baseline for video restoration with grouped spatial-temporal
  shift.
\newblock In \emph{Proceedings of the IEEE/CVF Conference on Computer Vision
  and Pattern Recognition}, pages 9822--9832, 2023.

\bibitem[Liang et~al.(2021)Liang, Cao, Sun, Zhang, Van~Gool, and
  Timofte]{liang2021swinir}
Jingyun Liang, Jiezhang Cao, Guolei Sun, Kai Zhang, Luc Van~Gool, and Radu
  Timofte.
\newblock Swinir: Image restoration using swin transformer.
\newblock In \emph{Proceedings of the IEEE/CVF international conference on
  computer vision}, pages 1833--1844, 2021.

\bibitem[Liang et~al.(2022)Liang, Fan, Xiang, Ranjan, Ilg, Green, Cao, Zhang,
  Timofte, and Gool]{liang2022recurrent}
Jingyun Liang, Yuchen Fan, Xiaoyu Xiang, Rakesh Ranjan, Eddy Ilg, Simon Green,
  Jiezhang Cao, Kai Zhang, Radu Timofte, and Luc~V Gool.
\newblock Recurrent video restoration transformer with guided deformable
  attention.
\newblock \emph{Advances in Neural Information Processing Systems},
  35:\penalty0 378--393, 2022.

\bibitem[Liang et~al.(2024)Liang, Cao, Fan, Zhang, Ranjan, Li, Timofte, and
  Van~Gool]{liang2024vrt}
Jingyun Liang, Jiezhang Cao, Yuchen Fan, Kai Zhang, Rakesh Ranjan, Yawei Li,
  Radu Timofte, and Luc Van~Gool.
\newblock Vrt: A video restoration transformer.
\newblock \emph{IEEE Transactions on Image Processing}, 2024.

\bibitem[Maggioni et~al.(2012)Maggioni, Boracchi, Foi, and
  Egiazarian]{maggioni2012bm4d}
Matteo Maggioni, Giacomo Boracchi, Alessandro Foi, and Karen Egiazarian.
\newblock Video denoising, deblocking, and enhancement through separable 4-d
  nonlocal spatiotemporal transforms.
\newblock \emph{IEEE Transactions on Image Processing}, 2012.

\bibitem[Maggioni et~al.(2021)Maggioni, Huang, Li, Xiao, Fu, and
  Song]{maggioni2021efficient}
Matteo Maggioni, Yibin Huang, Cheng Li, Shuai Xiao, Zhongqian Fu, and Fenglong
  Song.
\newblock Efficient multi-stage video denoising with recurrent spatio-temporal
  fusion.
\newblock In \emph{IEEE Conference on Computer Vision and Pattern Recognition},
  2021.

\bibitem[Nah et~al.(2018)Nah, Kim, and
  Lee]{nah2018deepmultiscaleconvolutionalneural}
Seungjun Nah, Tae~Hyun Kim, and Kyoung~Mu Lee.
\newblock Deep multi-scale convolutional neural network for dynamic scene
  deblurring, 2018.

\bibitem[Nah et~al.(2019)Nah, Baik, Hong, Moon, Son, Timofte, and
  Mu~Lee]{nah2019ntire}
Seungjun Nah, Sungyong Baik, Seokil Hong, Gyeongsik Moon, Sanghyun Son, Radu
  Timofte, and Kyoung Mu~Lee.
\newblock Ntire 2019 challenge on video deblurring and super-resolution:
  Dataset and study.
\newblock In \emph{Proceedings of the IEEE/CVF conference on computer vision
  and pattern recognition workshops}, pages 0--0, 2019.

\bibitem[Pan et~al.(2020)Pan, Bai, and Tang]{pan2020cascaded}
Jinshan Pan, Haoran Bai, and Jinhui Tang.
\newblock Cascaded deep video deblurring using temporal sharpness prior.
\newblock In \emph{Proceedings of the IEEE/CVF conference on computer vision
  and pattern recognition}, pages 3043--3051, 2020.

\bibitem[Ren et~al.(2017)Ren, Li, Liu, and Zeng]{8296826}
Zhihang Ren, Jiajia Li, Shuaicheng Liu, and Bing Zeng.
\newblock Meshflow video denoising.
\newblock In \emph{2017 IEEE International Conference on Image Processing
  (ICIP)}, pages 2966--2970, 2017.

\bibitem[Sheth et~al.(2021)Sheth, Mohan, Vincent, Manzorro, Crozier, Khapra,
  Simoncelli, and Fernandez-Granda]{sheth2021unsupervised}
Dev~Yashpal Sheth, Sreyas Mohan, Joshua~L Vincent, Ramon Manzorro, Peter~A
  Crozier, Mitesh~M Khapra, Eero~P Simoncelli, and Carlos Fernandez-Granda.
\newblock Unsupervised deep video denoising.
\newblock In \emph{Proceedings of the IEEE/CVF international conference on
  computer vision}, pages 1759--1768, 2021.

\bibitem[Son et~al.(2021)Son, Lee, Lee, Cho, and Lee]{son2021recurrent}
Hyeongseok Son, Junyong Lee, Jonghyeop Lee, Sunghyun Cho, and Seungyong Lee.
\newblock Recurrent video deblurring with blur-invariant motion estimation and
  pixel volumes.
\newblock \emph{ACM Transactions on Graphics (TOG)}, 40\penalty0 (5):\penalty0
  1--18, 2021.

\bibitem[Song et~al.(2022)Song, Zhang, and Ayd{\i}n]{song2022tempformer}
Mingyang Song, Yang Zhang, and Tun{\c{c}}~O Ayd{\i}n.
\newblock Tempformer: Temporally consistent transformer for video denoising.
\newblock In \emph{European conference on computer vision}, pages 481--496.
  Springer, 2022.

\bibitem[Staudemeyer and Morris(2019)]{staudemeyer2019understanding}
Ralf~C Staudemeyer and Eric~Rothstein Morris.
\newblock Understanding lstm--a tutorial into long short-term memory recurrent
  neural networks.
\newblock \emph{arXiv preprint arXiv:1909.09586}, 2019.

\bibitem[Su et~al.(2017)Su, Delbracio, Wang, Sapiro, Heidrich, and
  Wang]{su2017deep}
Shuochen Su, Mauricio Delbracio, Jue Wang, Guillermo Sapiro, Wolfgang Heidrich,
  and Oliver Wang.
\newblock Deep video deblurring for hand-held cameras.
\newblock In \emph{Proceedings of the IEEE conference on computer vision and
  pattern recognition}, pages 1279--1288, 2017.

\bibitem[Suin and Rajagopalan(2021)]{suin2021gated}
Maitreya Suin and AN Rajagopalan.
\newblock Gated spatio-temporal attention-guided video deblurring.
\newblock In \emph{Proceedings of the IEEE/CVF Conference on Computer Vision
  and Pattern Recognition}, pages 7802--7811, 2021.

\bibitem[Sun et~al.(2018)Sun, Yang, Liu, and Kautz]{sun2018pwc}
Deqing Sun, Xiaodong Yang, Ming-Yu Liu, and Jan Kautz.
\newblock Pwc-net: Cnns for optical flow using pyramid, warping, and cost
  volume.
\newblock In \emph{Proceedings of the IEEE conference on computer vision and
  pattern recognition}, pages 8934--8943, 2018.

\bibitem[Tassano et~al.(2019)Tassano, Delon, and Veit]{tassano2019dvdnet}
Matias Tassano, Julie Delon, and Thomas Veit.
\newblock Dvdnet: A fast network for deep video denoising.
\newblock In \emph{IEEE International Conference on Image Processing}, 2019.

\bibitem[Tassano et~al.(2020)Tassano, Delon, and Veit]{tassano2020fastdvdnet}
Matias Tassano, Julie Delon, and Thomas Veit.
\newblock Fastdvdnet: Towards real-time deep video denoising without flow
  estimation.
\newblock In \emph{IEEE Conference on Computer Vision and Pattern Recognition},
  2020.

\bibitem[Teed and Deng(2020)]{teed2020raft}
Zachary Teed and Jia Deng.
\newblock Raft: Recurrent all-pairs field transforms for optical flow.
\newblock In \emph{Computer Vision--ECCV 2020: 16th European Conference,
  Glasgow, UK, August 23--28, 2020, Proceedings, Part II 16}, pages 402--419.
  Springer, 2020.

\bibitem[Tian et~al.(2020)Tian, Zhang, Fu, and Xu]{tian2020tdan}
Yapeng Tian, Yulun Zhang, Yun Fu, and Chenliang Xu.
\newblock Tdan: Temporally-deformable alignment network for video
  super-resolution.
\newblock In \emph{Proceedings of the IEEE/CVF conference on computer vision
  and pattern recognition}, pages 3360--3369, 2020.

\bibitem[Truong et~al.(2024)Truong, Nguyen, Hua, and Yeung]{truong2024self}
Quang-Trung Truong, Duc~Thanh Nguyen, Binh-Son Hua, and Sai-Kit Yeung.
\newblock Self-supervised video object segmentation with distillation learning
  of deformable attention.
\newblock \emph{arXiv preprint arXiv:2401.13937}, 2024.

\bibitem[Vaksman et~al.(2021{\natexlab{a}})Vaksman, Elad, and
  Milanfar]{vaksman2021pacnet}
Gregory Vaksman, Michael Elad, and Peyman Milanfar.
\newblock Patch craft: Video denoising by deep modeling and patch matching.
\newblock In \emph{IEEE International Conference on Computer Vision},
  2021{\natexlab{a}}.

\bibitem[Vaksman et~al.(2021{\natexlab{b}})Vaksman, Elad, and
  Milanfar]{vaksman2021patch}
Gregory Vaksman, Michael Elad, and Peyman Milanfar.
\newblock Patch craft: Video denoising by deep modeling and patch matching.
\newblock In \emph{IEEE International Conference on Computer Vision},
  2021{\natexlab{b}}.

\bibitem[Vaswani et~al.(2017)Vaswani, Shazeer, Parmar, Uszkoreit, Jones, Gomez,
  Kaiser, and Polosukhin]{vaswani2017attention}
Ashish Vaswani, Noam Shazeer, Niki Parmar, Jakob Uszkoreit, Llion Jones,
  Aidan~N Gomez, {\L}ukasz Kaiser, and Illia Polosukhin.
\newblock Attention is all you need.
\newblock \emph{Advances in neural information processing systems}, 30, 2017.

\bibitem[Wang et~al.(2019)Wang, Chan, Yu, Dong, and Change~Loy]{wang2019edvr}
Xintao Wang, Kelvin~CK Chan, Ke Yu, Chao Dong, and Chen Change~Loy.
\newblock Edvr: Video restoration with enhanced deformable convolutional
  networks.
\newblock In \emph{IEEE Conference on Computer Vision and Pattern Recognition
  Workshops}, 2019.

\bibitem[Wang et~al.(2018)Wang, Yang, Yang, Zhao, Wang, and
  Xu]{wang2018occlusion}
Yang Wang, Yi Yang, Zhenheng Yang, Liang Zhao, Peng Wang, and Wei Xu.
\newblock Occlusion aware unsupervised learning of optical flow.
\newblock In \emph{Proceedings of the IEEE conference on computer vision and
  pattern recognition}, pages 4884--4893, 2018.

\bibitem[Wang et~al.(2023)Wang, Shi, Li, Huang, Cao, Zhang, Xian, and
  Lin]{wang2023neural}
Yiran Wang, Min Shi, Jiaqi Li, Zihao Huang, Zhiguo Cao, Jianming Zhang, Ke
  Xian, and Guosheng Lin.
\newblock Neural video depth stabilizer.
\newblock In \emph{Proceedings of the IEEE/CVF International Conference on
  Computer Vision}, pages 9466--9476, 2023.

\bibitem[Xia et~al.(2022)Xia, Pan, Song, Li, and Huang]{xia2022vision}
Zhuofan Xia, Xuran Pan, Shiji Song, Li~Erran Li, and Gao Huang.
\newblock Vision transformer with deformable attention.
\newblock In \emph{Proceedings of the IEEE/CVF conference on computer vision
  and pattern recognition}, pages 4794--4803, 2022.

\bibitem[Xian et~al.(2018)Xian, Shen, Cao, Lu, Xiao, Li, and
  Luo]{xian2018monocular}
Ke Xian, Chunhua Shen, Zhiguo Cao, Hao Lu, Yang Xiao, Ruibo Li, and Zhenbo Luo.
\newblock Monocular relative depth perception with web stereo data supervision.
\newblock In \emph{Proceedings of the IEEE Conference on Computer Vision and
  Pattern Recognition}, pages 311--320, 2018.

\bibitem[Xiang et~al.(2020)Xiang, Wei, and Pan]{xiang2020deep}
Xinguang Xiang, Hao Wei, and Jinshan Pan.
\newblock Deep video deblurring using sharpness features from exemplars.
\newblock \emph{IEEE Transactions on Image Processing}, 29:\penalty0
  8976--8987, 2020.

\bibitem[Xue et~al.(2019{\natexlab{a}})Xue, Chen, Wu, Wei, and
  Freeman]{xue2019toflow}
Tianfan Xue, Baian Chen, Jiajun Wu, Donglai Wei, and William~T Freeman.
\newblock Video enhancement with task-oriented flow.
\newblock \emph{International Journal of Computer Vision}, 127:\penalty0
  1106--1125, 2019{\natexlab{a}}.

\bibitem[Xue et~al.(2019{\natexlab{b}})Xue, Chen, Wu, Wei, and
  Freeman]{xue2019video}
Tianfan Xue, Baian Chen, Jiajun Wu, Donglai Wei, and William~T Freeman.
\newblock Video enhancement with task-oriented flow.
\newblock \emph{International Journal of Computer Vision}, 127:\penalty0
  1106--1125, 2019{\natexlab{b}}.

\bibitem[Yi et~al.(2019)Yi, Wang, Jiang, Jiang, and Ma]{yi2019progressive}
Peng Yi, Zhongyuan Wang, Kui Jiang, Junjun Jiang, and Jiayi Ma.
\newblock Progressive fusion video super-resolution network via exploiting
  non-local spatio-temporal correlations.
\newblock In \emph{Proceedings of the IEEE/CVF international conference on
  computer vision}, pages 3106--3115, 2019.

\bibitem[Yu et~al.(2019{\natexlab{a}})Yu, Park, and Jeong]{yu2019didn}
Songhyun Yu, Bumjun Park, and Jechang Jeong.
\newblock Deep iterative down-up cnn for image denoising.
\newblock In \emph{IEEE Conference on Computer Vision and Pattern Recognition
  workshops}, pages 0--0, 2019{\natexlab{a}}.

\bibitem[Yu et~al.(2020)Yu, Park, Park, and Jeong]{9150666}
Songhyun Yu, Bumjun Park, Junwoo Park, and Jechang Jeong.
\newblock Joint learning of blind video denoising and optical flow estimation.
\newblock In \emph{2020 IEEE/CVF Conference on Computer Vision and Pattern
  Recognition Workshops (CVPRW)}, pages 2099--2108, 2020.

\bibitem[Yu et~al.(2019{\natexlab{b}})Yu, Si, Hu, and Zhang]{yu2019review}
Yong Yu, Xiaosheng Si, Changhua Hu, and Jianxun Zhang.
\newblock A review of recurrent neural networks: Lstm cells and network
  architectures.
\newblock \emph{Neural computation}, 31\penalty0 (7):\penalty0 1235--1270,
  2019{\natexlab{b}}.

\bibitem[Yue et~al.(2020)Yue, Cao, Liao, Chu, and Yang]{yue2020supervised}
Huanjing Yue, Cong Cao, Lei Liao, Ronghe Chu, and Jingyu Yang.
\newblock Supervised raw video denoising with a benchmark dataset on dynamic
  scenes.
\newblock In \emph{IEEE Conference on Computer Vision and Pattern Recognition},
  2020.

\bibitem[Zhang et~al.(2024)Zhang, Xie, and Yao]{zhang2024blur}
Huicong Zhang, Haozhe Xie, and Hongxun Yao.
\newblock Blur-aware spatio-temporal sparse transformer for video deblurring.
\newblock In \emph{Proceedings of the IEEE/CVF Conference on Computer Vision
  and Pattern Recognition}, pages 2673--2681, 2024.

\bibitem[Zhang et~al.(2018)Zhang, Luo, Zhong, Ma, Liu, and
  Li]{zhang2018adversarial}
Kaihao Zhang, Wenhan Luo, Yiran Zhong, Lin Ma, Wei Liu, and Hongdong Li.
\newblock Adversarial spatio-temporal learning for video deblurring.
\newblock \emph{IEEE Transactions on Image Processing}, 28\penalty0
  (1):\penalty0 291--301, 2018.

\bibitem[Zhang et~al.(2022)Zhang, Li, Liang, Cao, Zhang, Tang, Timofte, and
  Van~Gool]{zhang2022scunet}
Kai Zhang, Yawei Li, Jingyun Liang, Jiezhang Cao, Yulun Zhang, Hao Tang, Radu
  Timofte, and Luc Van~Gool.
\newblock Practical blind denoising via swin-conv-unet and data synthesis.
\newblock \emph{arXiv preprint arXiv:2203.13278}, 2022.

\bibitem[Zhou et~al.(2019)Zhou, Zhang, Pan, Xie, Zuo, and Ren]{zhou2019spatio}
Shangchen Zhou, Jiawei Zhang, Jinshan Pan, Haozhe Xie, Wangmeng Zuo, and Jimmy
  Ren.
\newblock Spatio-temporal filter adaptive network for video deblurring.
\newblock In \emph{IEEE International Conference on Computer Vision}, pages
  2482--2491, 2019.

\bibitem[Zhu et~al.(2019)Zhu, Hu, Lin, and Dai]{zhu2019dcnv2}
Xizhou Zhu, Han Hu, Stephen Lin, and Jifeng Dai.
\newblock Deformable convnets v2: More deformable, better results.
\newblock In \emph{IEEE Conference on Computer Vision and Pattern Recognition},
  2019.

\end{thebibliography}
}

\end{document}